\documentclass[10pt,twocolumn,letterpaper]{article}

\usepackage{cvpr}              

\usepackage[T1]{fontenc}
\usepackage[utf8]{inputenc}
\usepackage{amsthm}
\usepackage{amsfonts}
\usepackage{bm}
\usepackage{commath}
\usepackage{dsfont}
\usepackage{pifont}
\usepackage{url}
\usepackage{mathtools}
\usepackage{dsfont}
\usepackage{mathrsfs}
\usepackage[short]{optidef}
\usepackage{stmaryrd}
\usepackage{nicefrac}
\usepackage{xcolor}
\usepackage[separate-uncertainty=true]{siunitx}
\usepackage[ruled, vlined, linesnumbered, noend]{algorithm2e}
\usepackage{microtype}
\usepackage{cases}
\usepackage{import}
\usepackage{transparent}
\usepackage{multirow}
\usepackage{multido}
\usepackage{makecell}
\usepackage{graphics}
\usepackage{wrapfig}
\usepackage{comment}
\usepackage{datetime}
\usepackage{float}
\usepackage{bbm}

\definecolor{cvprblue}{rgb}{0.21,0.49,0.74}
\usepackage[pagebackref,breaklinks,colorlinks,citecolor=cvprblue]{hyperref}

\def\paperID{} 
\def\confName{CVPR}
\def\confYear{2024}
\def\*#1{\mathbf{#1}}

\usepackage{adjustbox}
\newcolumntype{R}[2]{%
    >{\adjustbox{angle=#1,lap=\width-(#2)}\bgroup}%
    l%
    <{\egroup}%
}
\newcommand*\rot{\multicolumn{1}{R{45}{1em}}}
\newcolumntype{P}[2]{%
    >{\adjustbox{angle=#1,lap=\width-(#2)}\bgroup}%
    c%
    <{\egroup}%
}

\title{Exploring 3D-aware Latent Spaces for Efficiently Learning Numerous Scenes}

\def\authorA{Karim Kassab}

\def\authorB{Antoine Schnepf}

\def\authorC{Jean-Yves Franceschi}

\def\authorD{Laurent Caraffa}

\def\authorE{Jeremie Mary}

\def\authorF{Valérie Gouet-Brunet}

\def\authorG{Flavian Vasile}

\def\authorH{Andrew Comport}

\author{
\large \authorB \textsuperscript{~*~1,3} \hspace{0.5cm}
\large \authorA \textsuperscript{*~1,2} \hspace{0.5cm}
\large \authorC \textsuperscript{1} \hspace{0.5cm}
\large \authorD \textsuperscript{2} \hspace{0.5cm} \vspace{0.1cm}\\
\large \authorG \textsuperscript{1} \hspace{0.5cm}
\large \authorE \textsuperscript{1} \hspace{0.5cm}
\large \authorH \textsuperscript{3} \hspace{0.5cm}
\large \authorF \textsuperscript{2} \vspace{0.3cm}\\
\small \textsuperscript{*} Equal contributions \\
\small \textsuperscript{1} Criteo AI Lab, Paris, France \\
\small \textsuperscript{2} LASTIG, Université Gustave Eiffel, IGN-ENSG, F-94160 Saint-Mandé \\
\small \textsuperscript{3} Université Côte d’Azur, CNRS, I3S, France
}

\newcommand{\cmark}{\textcolor{green}{\ding{51}}}%
\newcommand{\xmark}{\textcolor{red}{\ding{55}}}%
\usepackage{tablefootnote}

\SetCommentSty{mycommfont}

\begin{document}
\maketitle

\begin{abstract}
We present a method enabling the scaling of NeRFs to learn a large number of semantically-similar scenes. 
We combine two techniques to improve the required training time and memory cost per scene.
First, we learn a 3D-aware latent space in which we train Tri-Plane scene representations, hence reducing the resolution at which scenes are learned. 
Moreover, we present a way to share common information across scenes, hence allowing for a reduction of model complexity to learn a particular scene. 
Our method reduces effective per-scene memory costs by 44\% and per-scene time costs by 86\% when training 1000 scenes.
Our project page can be found at \url{https://3da-ae.github.io}~.
\end{abstract}

\section{Introduction}
\label{sec:intro}
The inverse graphics problem has proven to be a challenging quest in the domain of Computer Vision. 
While many methods have historically emerged \citep{Lumigraph,deepsdf,differentiable-rendering,nerf}, the question has mostly remained unchanged: how to model an object or scene, using only its captured images? 
While this question continues to be an active area of research, our work targets a scaled 
version of the original problem: how to model abundantly many objects at once?
This context is dissimilar to independently learning scenes, as one could exploit the inherent nature of the problem --- learning \emph{many} scenes --- to mitigate per-scene optimization costs.
Although modeling 3D objects and scenes from captured images has seen various applications (e.g.\ Virtual Reality, Robotics, Autonomous Navigation), scaling this problem to large amounts of objects unlocks new ways in which 3D modeling techniques could be leveraged (e.g.\ modeling an inventory of products for commerce, integrating real artifacts in virtual spaces).

In this paper, we introduce a novel technique enabling the scaling of the inverse graphics problem.
To do so, our work aims to compress scene representations to model only essential information within a particular scene.
To achieve this, we build on two main ideas: we learn a \textbf{3D-aware (2D) latent space} in which we train our scene representations, and we introduce \textbf{cross-scene feature sharing} to avoid redundantly learning similar information.
\begin{figure}[t]
    \centering
    \includegraphics[width=\columnwidth]{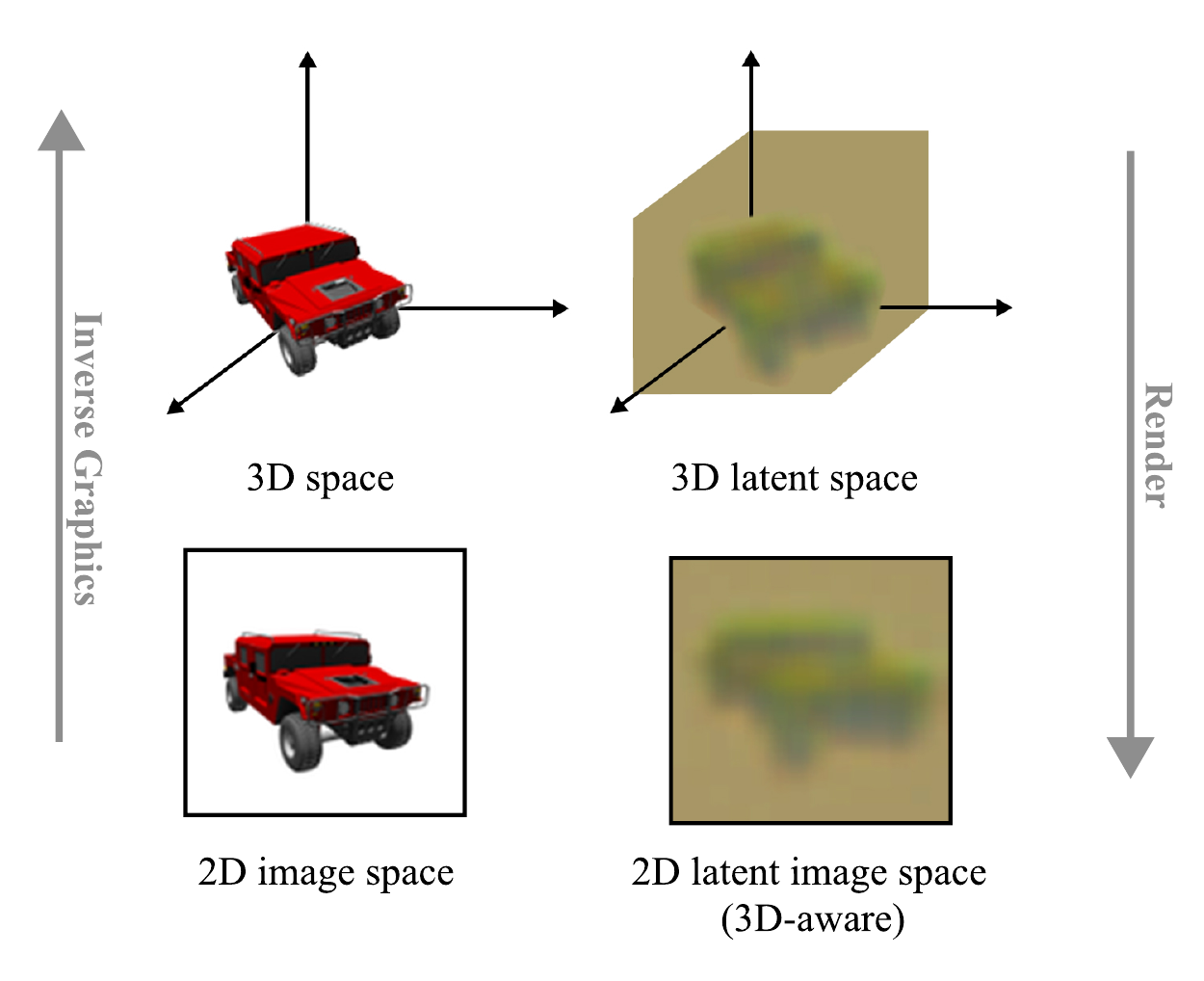}
    \caption{\textbf{3D-aware latent space.}  We draw inspiration from the relationship between the 3D space and image space and introduce the idea of a 3D latent space. We propose a 3D-aware autoencoder that encodes images into a 3D-aware (2D) latent image space, in which we train our scene representations.}
    \label{fig:3D-latent-space}
\end{figure}

\textbf{Latent spaces} enable the representation of high-resolution RGB images in a compact form, through the process of encoding and decoding.
Particularly, Auto-Encoders (AEs) learn a lower-dimensional latent space able to capture the underlying structure and diversity of a dataset of images.
However, as their optimization is devoid of geometrical constraints, latent spaces generally lack 3D structure, which is particularly crucial in 3D applications as this promotes a discrepancy between the latent space and the real world images it represents.
One particular downfall of this discrepancy is the absence of 3D-consistency between two latent images encoding two 3D-consistent images.
In this work, we present a novel approach to regularize a latent space with geometric constraints by leveraging Tri-Planes representations \citep{eg3d}.
This not only adds 3D consistency to the latent space, but also enables the use of recent neural rendering techniques within this latent space, which accelerates their optimization and rendering times, and lowers their memory footprint. 
\cref{fig:3D-latent-space} illustrates the intuition behind our 3D latent space, and the analogy with the natural 3D space.

To go further, we also minimize the amount of information we learn per scene representation by introducing the notion of \textbf{globally shared representations}.
Here, we train $M$ representations that are shared among scenes, and that learn \emph{global} information about the scenes within the dataset at hand.
This avoids learning redundant information across scenes and hence  minimizes both the per-scene compute time and memory cost when learning numerous scenes.

A summary of our contribution can be found below:
\begin{itemize}
    \item We build a 3D-aware latent space in which neural scene representations can be trained,
    \item We present an approach to minimize the capacity needed to model a latent scene by sharing common globally-trained scene representations across scenes,
    \item Our work can learn 1000 scenes with  86\% less time and 44\% less memory than our base representation.
\end{itemize}

\section{Related Work}
\label{sec:related-work}
\paragraph{NeRF resource reductions. }
Neural Radiance Fields \citep[NeRF]{nerf} achieve impressive performances on the task of Novel View Synthesis (NVS) by adopting a purely implicit representation to model scenes. 
On one end of the spectrum, some NeRF methods \citep{mipnerf,mipnerf-360} achieve exceptional quality while requiring low memory capacity to store scenes, as they represent scenes through the weights of neural networks.
This however comes with the sacrifice of high training and rendering times due to bottlenecks in volume rendering.
To alleviate these issues, some representations trade-off compute time for memory usage by explicitly storing proxy features for the emitted radiances and densities in a 3D data structure (e.g.\  voxel based representations \citep{direct-voxel-grid-opt,tensorf,plenoctrees,instantngp} or plane-based representations \citep{eg3d,kplanes,hexplane}). 
This allows for the use of a significantly smaller neural network as compared to purely implicit representations.
On the other end of the spectrum, \citet{gaussian-splatting,plenoxels} completely forgo the use of neural networks, achieving real-time rendering, but at high memory costs.

In contrast to previous works, we propose a method orthogonal to the aforementioned spectrum and sidestepping the time-memory trade-off; by accelerating scene learning all the while lowering the memory footprint of individual scenes.
This is partly done by training our scene representations on a compressed version of training images. 
To do so, we present a novel 3D-aware latent feature space. 

\paragraph{Neural Feature Fields. }
Neural Feature Fields extend Neural Radiance Fields to render feature images instead of color images, while using the same volume rendering equations.
First methods exploring feature fields propose the 3D distillation of features by jointly training a radiance field and a feature field.
This allows for modeling an \emph{individual} scene in a feature field rather than a radiance field.
This is particularly interesting, as this representation unlocks many subsequent applications, like 3D object detection and segmentation \citep{feature-fields}, 3D editing \citep{decomposing-nerf,control-nerf} and semantic understanding of scenes \citep{clip-fields, lerf}.
\citet{latentnerf,ditto-nerf} train a feature field that renders in the latent space of Stable Diffusion \citep{stable-diffusion}. These feature fields are optimized such that their renderings match the posterior distribution of Stable Diffusion under a descriptive text prompt, hence enabling text-to-3D generation.
\citet{featurenerf,pixelnerf,gen-nvs} use a pre-trained encoder to generate feature fields from single images, hence enabling single-image-to-3D generation.
\citet{decode-nerf} are closest to our work, as they model individual scenes in a latent feature space with a decoder neural network, thus accelerating volume rendering thanks to the reduced latent-space dimension. However, this training is individually done for each scene, and no common latent space is learned across scenes.

Our work expands \citep{decode-nerf} in a different scope, more particularly by leveraging both an encoder and a decoder to learn a \emph{common} 3D-consistent latent space in which numerous scene representations can be trained.

\begin{figure*}[t]
    \centering
\includegraphics[width=0.95\textwidth]{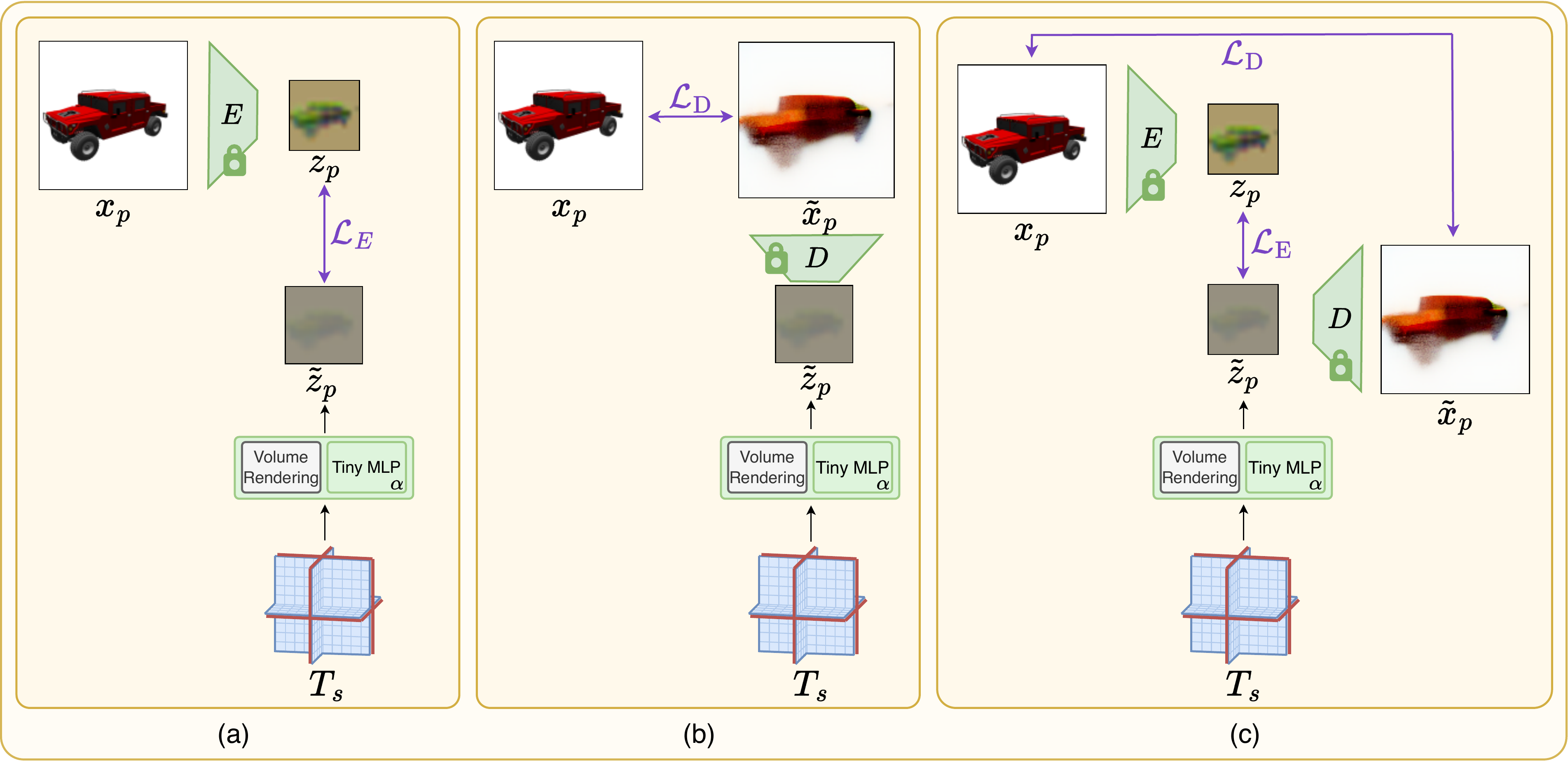}
    \caption{\textbf{Methods for learning scenes in a 3D-aware latent space.} Diagrams for (a) Encode-Scene, (b) Decode-Scene, and (c) Encode-Decode-Scene, the proposed methods to train Tri-Plane scene representations in a 3D-aware latent space.}
    \label{fig:encode_decode_nerf}
\end{figure*}
\paragraph{Meta-Learning base networks for INRs. } 
To reduce model complexity, previous works have explored learning and modulating shared base networks to represent common structure within a set of signals modeled by Implicit Neural Representations (INRs).
More particularly, to create a \emph{functaset} of NeRFs, \citet{data-to-functa} learn a shared base network thanks to which NeRFs can be trained by only optimizing modulations of the base network.
Our work draws inspiration from this and introduces global Tri-Planes, which consist of shared scene representations storing common information and structure across a set of scenes.
This enables in our case the reduction of per-scene model complexity.
Previous works have also shown that meta-learning such shared networks enables fitting Neural Radiance Fields and Signed Distance Functions in only a few optimization steps \citep{learned-inits,meta-sdf}.
More generally, previous works have also shown the advantage of learning priors over a subset of scenes. \citep{nerf-vae, laser, data-to-functa, deepsdf}.
Our work parallels this by pre-training parts of our pipeline on the training set, hence learning a prior over the scene distribution before applying it on new scenes.

\section{Method}
\label{sec:method}
In this section, we present the main components of our method and elucidate the intuition behind our choices. 
We start by presenting three methods with which one could train a scene representation in a 3D-aware latent space: \textbf{Encode-Scene}, \textbf{Decode-Scene}, and \textbf{Encode-Decode-Scene}.
We subsequently present our 3D-aware Autoencoder (\textbf{3Da-AE}) which builds upon these methods to learn a 3D-aware latent space.
We also present \textbf{Micro-Macro representations}, which constitute an additional component in our pipeline that further accelerates training while lowering its memory footprint thanks to information sharing.
Finally, we show how to use 3Da-AE and Micro-Macro Tri-Planes to solve the problem of learning numerous scenes.

\subsection{Prerequisites}

\paragraph{Autoencoder.}
An autoencoder (AE) is a compression model trained to learn efficient, low-resolution representations of images.
To achieve this, the model is trained via a reconstruction loss where images are passed through an information bottleneck:
\begin{equation}
\begin{split}
    z &= E_{\psi}(x)~,\\
    \hat{x} &= D_{\phi}(z)~,\\
    \mathcal{L}_{\mathrm{ae}}{(\psi, \phi)} &= \mathbb{E}_x \Vert x - \hat{x} \rVert_2^2~,
\end{split}
\end{equation}
where $x$ is an image, $E_{\psi}$ and $D_{\phi}$ respectively represent the encoder and decoder with trainable parameters $\psi$ and $\phi$, and $z$ has a lower resolution than $x$.
In this work, we use the autoencoder of Stable Diffusion \citep{stable-diffusion} as a baseline. It works with a large range of input image resolutions, while reducing the resolution by a factor of 64 in the latent space.

\paragraph{3D consistency. }
The notion of 3D consistency refers to the underlying 3D geometry of 2D images. 
Formally, 3D consistency involves ensuring that corresponding points or features in different images represent the same physical point or object in the scene, despite variations in viewpoint, lighting or occlusion.
Note that while 3D consistency is natural for a set of posed images $\mathcal{X}_s = \{p, x_p\}_{p \in \mathcal{P}}$ obtained from a scene $s$ in the image space, it does not naturally extend to the latent space, as latent representations of two 3D-consistent images are not necessarily 3D consistent (\cref{fig:encodings_comparison}, bottom row).

\paragraph{Tri-Plane representation. }
Tri-Plane representations \citep{eg3d} are explicit-implicit scene representations enabling scene modeling in three axis-aligned orthogonal feature planes, each of resolution $T \times T$ with feature dimension $F$.
To query a 3D point $x \in \mathbb{R}^3$, it is projected onto each of the three planes to retrieve bilineraly interpolated feature vectors $F_{xy}$, $F_{xz}$ and $F_{yz}$.
These feature vectors are then aggregated via summation and passed into a small neural network to retrieve the corresponding color and density, which are then used for volume rendering \citep{volume-rendering}.
We adopt Tri-Plane representations for their relatively fast training times, as well as their explicit nature enabling their modularity, an essential property for our Micro-Macro decomposition (\cref{sec:Micro-Macro}).

\subsection{Latent NeRFs}
\label{sec:latent-nerfs}
We define a latent scene representation similarly to a classical scene representation, except that it is trained on latent encodings of the scene images.
In this section, we assume the existence of a 3D-aware latent space, and present three approaches to learn latent scene representations: \textbf{Encode-Scene}, \textbf{Decode-Scene}, and \textbf{Encode-Decode-Scene}, respectively utilizing the encoder, decoder and both modules of a 3D-aware autoencoder.
\cref{fig:encode_decode_nerf} illustrates these methods.

\subsubsection{Encode-Scene}
\label{sec:Encode-LN}
Encode-Scene takes the most similar approach compared to classical NeRF training in order to train NeRFs in a latent space.
It is a two-step process where latent images are first obtained from training images using the encoder, and are then used to train the latent NeRF with the usual NeRF photometric loss:
\begin{equation}
    \min_{\alpha,T}~\mathcal{L}_E(\alpha,T) \triangleq \mathbb{E}_{x_p} \lVert E(x_p) - \mathcal{R}_\alpha(T, p) \rVert~,
\end{equation}
where $E(x_p)$ is the encoding of an image $x_p$ with camera pose $p$, and $\mathcal{R}_\alpha(T, p)$ represents the rendered image from the Tri-Plane $T$ queried at pose $p$.
Note that the main advantage of this approach is its accelerated training and rendering procedures, as all the training images can be encoded into their latent representations once, and then cached. 
As these latent representations are 64 times smaller in resolution, the rendering algorithm has to query 64 times less pixels, which greatly accelerates rendering and hence the training.

\begin{figure}[t]
    \centering
    \includegraphics[width=1\columnwidth]{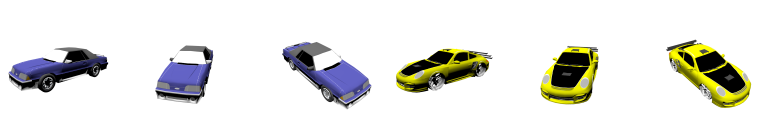}
    \includegraphics[width=1\columnwidth]{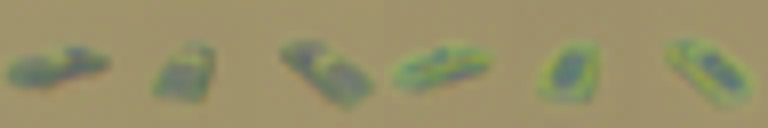}
    \includegraphics[width=1\columnwidth]{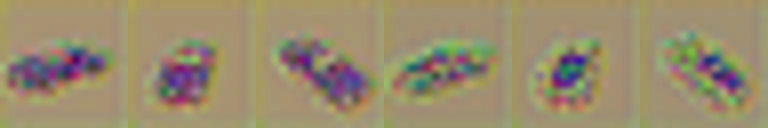}
    \caption{\textbf{Latent space comparison. }  Top: ground truth image. Middle: latent image obtained with the 3D-aware encoder. Bottom: latent image obtained with the baseline encoder. Qualitative results show that our 3D-aware encoder better preserves 3D consistency and geometry in the latent space.}
    \label{fig:encodings_comparison}
\end{figure}
\subsubsection{Decode-Scene}
\label{sec:Decode-LN}
Decode-Scene takes a different approach to train NeRFs in the latent space. 
Here, we use the decoder and supervise the pipeline with RGB images, all while keeping the NeRF in the latent space. 
More particularly, the NeRF here is tasked to find a 3D-consistent latent object that, when rendered into latent images, decodes to its corresponding RGB images. 
Hence, the NeRF here is supervised via a photometric loss computed in the RGB space:
\begin{equation}
    \min_{\alpha,T}~\mathcal{L}_D(\alpha,T) \triangleq \mathbb{E}_{x_p} \lVert x_p - D(\mathcal{R}_\alpha(T, p)) \rVert~,
\end{equation}
where $D$ represents the decoder.
While rendering is still applied in the latent space in Decode-Scene, this is generally a slower method compared to Encode-Scene. This is because no latent images can be cached, and a gradient step requires differentiation through all the parameters of the decoder. 

\begin{figure}[t]
    \centering
    \begin{subfigure}{0.495\columnwidth}
        \centering
        \includegraphics[width=\textwidth]{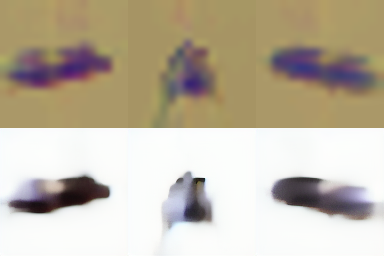}
        \label{fig:standard-VAE}
        \caption{Standard AE}
    \end{subfigure}
    \begin{subfigure}{0.495\columnwidth}
        \centering
        \includegraphics[width=\textwidth]{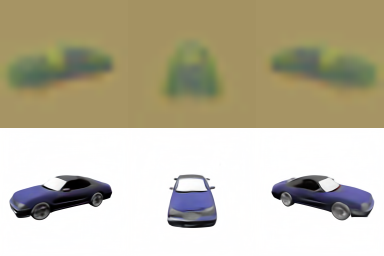}
        \label{fig:3D-aware-VAE}
        \caption{3D-aware AE}
    \end{subfigure}
    \caption{\textbf{Latent scenes comparison}. Visualization of Tri-Planes renderings and their corresponding decodings after learning scenes in the latent space of a standard AE and our 3D-aware AE. All Tri-Planes are trained using the Encode-Scene pipeline.}
    \label{fig:qualitative-visuals}
\end{figure}
\begin{figure*}[ht]
    \centering
    \includegraphics[width=0.9\textwidth]{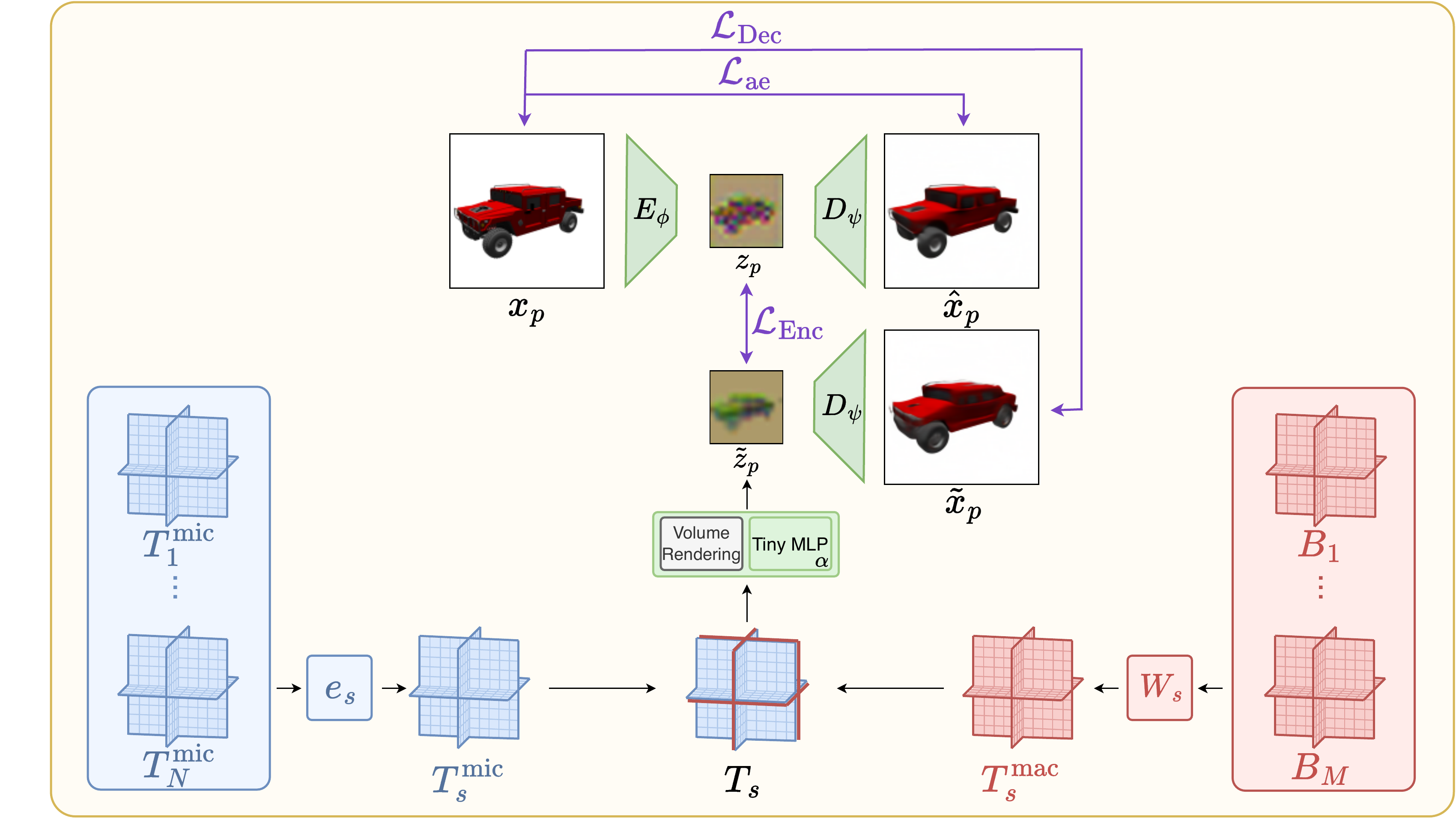}
    \caption{\textbf{3Da-AE training. } 
        We learn a 3D-aware latent space by regularizing its training with 3D constraints.
        To this end, we jointly train the encoder $E_\phi$, the decoder $D_\psi$ and $N$ scenes in this latent space. For each scene $s$, we learn a Tri-Planes representation $T_{s}$, built from the concatenation of local Tri-Planes $T_s^{mic}$ and global Tri-Planes $T_s^{mac}$.
        $T_s^{mic}$ is retrieved via a one-hot vector $e_s$ from a set of scene-specific planes stored in memory.
        $T_s^{mac}$ is computed from a summation of $M$ globally shared Tri-Planes, weighted with weights $W_s$.
    }
    \label{fig:schema}
\end{figure*}
\subsubsection{Encode-Decode-Scene}
\label{sec:ED-LN}
Encode-Decode-Scene is a mixture of the former approaches, where both $\mathcal{L}_E$ and $\mathcal{L}_D$ are used to train the latent NeRF:
\begin{equation}
    \min_{\alpha,T}~(1-t) \mathcal{L}_D(\alpha,T) + t \mathcal{L}_E(\alpha,T)~,
\end{equation}
where $t \in (0,1)$. 
This method takes the best of both worlds by supervising both the rendered latent and the decoded image to train a scene representation, ensuring good latent scene representations and high-fidelity color images.

Qualitative results obtained when training Tri-Planes with the baseline AE and our 3D-aware AE are illustrated in \cref{fig:qualitative-visuals}.
While learning scenes in a standard latent space is feasible, it results in poor scene quality, as the latent images used for inverse graphics are not 3D-consistent (\cref{fig:encodings_comparison}, bottom row).

\subsection{3D-aware AE for a 3D-aware latent space}
\label{sec:3Da-VAE}
As presented, NeRFs trained in a standard latent space suffer from its 3D inconsistencies. 
To fix this issue, we fine-tune the autoencoder of Stable Diffusion \citep{stable-diffusion} to obtain a \mbox{3D-aware} Autoencoder (3Da-AE).  

To enforce 3D consistency in the latent space, we regularize it by training NeRFs on its latent images.
Intuitively, this encourages the encoder and the decoder to preserve 3D consistency, thanks to the NeRF's innate 3D-consistent nature.
To do so, we present a joint training process where we optimize $N_\mathrm{train}$ Tri-Planes modeling latent scenes, as well as the encoder and decoder.
\cref{fig:encodings_comparison} (middle row) illustrates the preservation of consistency of our method.

To achieve this joint training, we implement three losses, inspired from \cref{sec:latent-nerfs}, in order to simultaneously train the encoder, the decoder, as well as the latent NeRFs.
First, an autoencoder preservation loss $\mathcal{L}_\mathrm{ae}$ ensures the correct reconstruction of images after encoding and decoding. 
Second, $\mathcal{L}_\mathrm{Enc}$ reassembles $\mathcal{L}_E$ from Encode-Scene, but additionally supervises the encoder so that it produces 3D consistent latent images.
Third, $\mathcal{L}_\mathrm{Dec}$ reassembles $\mathcal{L}_D$ from Decode-Scene, but additionally trains the decoder to ensure that two 3D-consistent latent images are decoded into their corresponding 3D-consistent RGB images. 
Hence, the training objective for the 3Da-AE is written as:
\begin{equation}
\begin{split}
    &\begin{split}
        \min_{\phi, \psi,\alpha,T}~\lambda_\mathrm{ae}\mathcal{L}_\mathrm{ae} (\phi,\psi)+\lambda_\mathrm{Enc}\mathcal{L}&_\mathrm{Enc} (\phi,\alpha,T)\\
        & + \lambda_\mathrm{Dec}\mathcal{L}_\mathrm{Dec} ( \psi,\alpha,T)~,
    \end{split}\\
    &\text{with }\begin{cases}    
    &\mathcal{L}_\mathrm{ae} (\phi,\psi) = \mathbb{E}_{x_p} \lVert x_p - D_\psi(E_\phi(x_p)) \rVert~,\\
    &\mathcal{L}_\mathrm{Enc} (\phi,\alpha,T) = \mathbb{E}_{x_p} \lVert E_\phi (x_p) - \mathcal{R}_\alpha(T, p) \rVert~,\\
    &\mathcal{L}_\mathrm{Dec} (\psi,\alpha,T) = \mathbb{E}_{x_p} \lVert x_p - D_\psi (\mathcal{R}_\alpha(T, p)) \rVert~,
    \end{cases}
\end{split}
\end{equation}
where $\phi$, $\psi$ and $\alpha$ are shared parameters among all scenes.
$T$ consists of Tri-Plane parameters which we divide into scene-specific local parameters and globally-shared computed parameters.
For an overview of the full 3Da-AE pipeline, we refer the reader to \cref{fig:schema}.

\subsection{Micro-Macro Tri-Plane Decomposition}
\label{sec:Micro-Macro}
In this section, we present an additional approach to reduce the capacity needed to learn individual scenes through Tri-Plane scene representations.
As most of our scenes share similar structure, we sidestep repeatedly learning redundant information across scenes by integrating globally shared information into our scene representations, and modulating this shared information via scene-dependent weights.
While learning scenes in the latent space achieves complexity reductions through minimizing spatial resolutions, we aim to achieve this here by decomposing a scene representation into ``globally'' shared information and ``locally'' learned features.

Formally, we decompose a Tri-Plane representation $T_s$ modeling a scene $s$ into a locally trained Tri-Plane representation $T^\mathrm{mic}_s$ and a globally learned representation $T^\mathrm{mac}_s$:
\begin{equation}
    T_s = T^\mathrm{mic}_s \oplus T^\mathrm{mac}_s~,
\end{equation}
where $\oplus$ concatenates two Tri-Planes along the feature dimension.
We denote by $F^\mathrm{mic}$ the number of local feature in $T^\mathrm{mic}_s$  and by $F^\mathrm{mac}$ the number of global feature in $T^\mathrm{mac}_s$, with the total number of feature $F = F^\mathrm{mic} + F^\mathrm{mac}$. 

For $T^\mathrm{mac}_s$ to represent globally captured information, it is computed for each scene from globally learned Tri-Plane representations $\{B_i\}_{i=1}^M$ by the means of a weighted sum:
\begin{equation} \label{eq:micmac_decomposition}
    T^\mathrm{mac}_s = W_s B = \sum_{i=1}^{M} w_s^i B_i~,
\end{equation}
where $W_s$ are learned coefficients for scene $s$, and $B_i$ are jointly trained with every scene.
This approach accelerates our method and reduces its memory footprint, as we asymptotically reduce the number of trainable features by a factor of $\frac{F^{\mathrm{mac}}}{F^{\mathrm{mic}} + F^{\mathrm{mac}}}$~.

\subsection{Scaling 3Da-AE}

In order to leverage our 3Da-AE pipeline to train a large number of NeRFs, we utilize the ideas presented in \cref{sec:3Da-VAE,sec:Micro-Macro} within two stages: ``\textbf{Training 3Da-AE}'' (\cref{sec:training-3da-ae}) and ``\textbf{Exploiting 3Da-AE}'' (\cref{sec:exploiting-3Da-AE}).
The objective of the first stage is to train 3Da-AE and the global planes in order to obtain our 3D-aware latent space.
The goal in the second stage is to learn numerous scenes by using the 3D-aware latent space obtained in the first step, as well as the global planes.
We detail the two stages in the following sections.

\begin{algorithm}[t]
\DontPrintSemicolon
\caption{3Da-AE Training.}
\label{alg:training}
{\LinesNotNumbered
\KwIn{$\mathcal{X_{\mathrm{train}}}$, $E_\phi$, $D_\psi$, $\mathcal{R}_\alpha$, $N$, $\lambda_\mathrm{ae}$, $\lambda_\mathrm{Enc}$, $\lambda_\mathrm{Dec}$, \text{optimizer}}
\textbf{Random initialization:} $T^\mathrm{mic}, W, T^\mathrm{mac}$\;}

\For{$N$ steps} {
    \For{$\{ s, p, x_p \}$ in $\mathrm{shuffle}(\mathcal{X_{\mathrm{train}}})$} {
        \tcp{Compute local planes}
        $T^\mathrm{mic}_s~,~T^\mathrm{mac}_s \gets e_s T^\mathrm{mic}~,~W_s B$\;
        $T_s \gets T^\mathrm{mic}_s \oplus T^\mathrm{mac}_s$\;
        \tcp{Encode, Decode, Render}
        $z_p \gets E_\phi (x_p)$\;
        $\hat{x}_p \gets D_\psi (z_p) $\;
        $\tilde{z}_p \gets \mathcal{R}_\alpha(T_s, p)$\;
        $\tilde{x}_p \gets D_\psi (\tilde{z}_p)$\;
        \tcp{Compute losses and optimize}
        $\mathcal{L}_\mathrm{ae} \gets \lVert x_p - \hat{x}_p \rVert_2^2$\;
        $\mathcal{L}_\mathrm{Enc} \gets \lVert z_p - \tilde{z}_p \rVert_2^2$\;
        $\mathcal{L}_\mathrm{Dec} \gets \lVert x_p - \tilde{x}_p \rVert_2^2$\;
        $\mathcal{L} \gets \lambda_\mathrm{ae}\mathcal{L}_\mathrm{ae} + \lambda_\mathrm{Enc}\mathcal{L}_\mathrm{Enc} + \lambda_\mathrm{Dec}\mathcal{L}_\mathrm{Dec}$\;
        $T^{mic}_s, W_s, B, \alpha, \phi, \psi \gets \text{optimizer.step}(\mathcal{L})$\;
    }
}
\end{algorithm}

\subsubsection{Training 3Da-AE}
\label{sec:training-3da-ae}
In this stage, we learn a 3D-aware autoencoder.
Our training dataset $\mathcal{X}_{\mathrm{train}} = \{(s, p, x_p) \}_{s \in \mathcal{S},~p \in \mathcal{P}}$, where $\mathcal{S}$ denotes the set of scene indices and $\mathcal{P}$ the set of poses, is composed of $N_\mathrm{train}$ scenes from ShapeNet \citep{shapenet}.
We initialize the local Tri-Planes $\{ T_s^\mathrm{mic}\}_{s \in \mathcal{S}}$, as well as our global Tri-Planes $\{B_i\}_{i=1}^{M}$ and the scene-specific coefficients $\{W_s\}_{s \in \mathcal{S}}$. For each scene $s$, we compute the corresponding representation $T_s$ with the Micro-Macro decomposition (\cref{eq:micmac_decomposition}).

Given a posed image $(p, x_p)$ of a scene $s$, we use the autoencoder to obtain the latent image $z_p$ and reconstructed image $\hat{x}_p$. Subsequently, we render the triplane $T_s$ from pose $p$ to obtain the rendered latent $\tilde{z}_p$, which we decode to obtain $\tilde{x}_p$. The losses $\mathcal{L}_\mathrm{ae}$, $\mathcal{L}_\mathrm{Dec}$ and $\mathcal{L}_\mathrm{Enc}$ are estimated on a mini-batch and used to optimize the encoder, decoder, local Tri-Planes, global Tri-Planes and the scene-specific coefficients.
\cref{alg:training} details our training procedure.

Note that in practice, we begin this training with a warmup stage, during which only the local triplanes, global triplanes, and scene-specific coefficients are optimized. Here, we keep the autoencoder frozen as random gradients would back-propagate into $E_\phi$ and $D_\psi$.

\begin{algorithm}[t]
\DontPrintSemicolon
\caption{3Da-AE Exploitation.}
\label{alg:exploiting}
{\LinesNotNumbered
\KwIn{$\mathcal{X_{\mathrm{exploit}}}$, $E_\phi$, $D_\psi$, $B$, $\mathcal{R}_\alpha$, $N_1$, $N_2$, \text{optimizer}}
\textbf{Random initialization:} $T^\mathrm{mic}, W$\;}

\tcp{Encode-Scene}
\For{$N_1$ steps} {
    \For{$\{ s, p, x_p \}$ in $\mathrm{shuffle}(\mathcal{X_{\mathrm{exploit}}})$} {
    
        \tcp{Compute local planes}
        $T^\mathrm{mic}_s~,~T^\mathrm{mac}_s \gets e_s T^\mathrm{mic}~,~W_s B$\;
        $T_s \gets T^\mathrm{mic}_s \oplus T^\mathrm{mac}_s$\;
        
        \tcp{Encode, Render}
        $z_p \gets E_\phi (x_p)$\;
        $\tilde{z}_p \gets \mathcal{R}_\alpha(T_s, p)$\;

        \tcp{Compute losses and optimize}
        $\mathcal{L}_\mathrm{E} \gets \lVert z_p - \tilde{z}_p \rVert_2^2$\;
        $T^{mic}_s, W_s, B, \alpha \gets \text{optimizer.step}(\mathcal{L}_\mathrm{E})$\;
    }
}
\tcp{Decoder finetuning}
\For{$N_2$ steps} {
    \For{$\{ s, p, x_p \}$ in $\mathrm{shuffle}(\mathcal{X_{\mathrm{exploit}}})$} {

        \tcp{Compute local planes}
        $T^\mathrm{mic}_s~,~T^\mathrm{mac}_s \gets e_s T^\mathrm{mic}~,~W_s B$\;
        $T_s \gets T^\mathrm{mic}_s \oplus T^\mathrm{mac}_s$\;

        \tcp{Encode, Decode, Render}
        $\tilde{z}_p \gets \mathcal{R}_\alpha(T_s, p)$\;
        $\tilde{x}_p \gets D_\psi (\tilde{z}_p)$\;
        
        \tcp{Compute losses and optimize}
        $\mathcal{L}_\mathrm{Dec} \gets \lVert x_p - \tilde{x}_p \rVert_2^2$\;
        $T^{mic}_s, W_s, B, \alpha, \psi \gets \text{optimizer.step}(\mathcal{L}_\mathrm{Dec})$\;
    }
}
\end{algorithm}

\subsubsection{Exploiting 3Da-AE}
\begin{table*}[ht]
    \centering
    \resizebox{0.7\textwidth}{!}{
    \begin{tabular}{l c c c c c c}
        \toprule
        & \multicolumn{1}{p{1cm}}{\centering $t_\mathrm{scene}$ \\ (min)} 
        & \multicolumn{1}{p{1cm}}{\centering $t_\mathrm{scene}^\mathrm{eff}$ \\ (min)} 
        & \multicolumn{1}{p{1cm}}{\centering $m_\mathrm{scene}$ \\ (MB)} 
        & \multicolumn{1}{p{1cm}}{\centering $m_\mathrm{scene}^\mathrm{eff}$ \\ (MB)} 
        & \multicolumn{1}{p{2cm}}{\centering Rendering \\ Time (ms)} 
        & \multicolumn{1}{p{2cm}}{\centering Rendering \\ Resolution} \\ \midrule
        Encoder & --- & --- & 0 & 0.13 & --- & --- \\
        Decoder & --- & --- & 0 & 0.19 & 9.7 & $128 \times 128$ \\ \midrule
        Tri-Planes (RGB) & 32 & 32 & 1.5 & 1.5 & 23.3 & $128 \times 128$ \\
        Our method & 2 & 4.5 & 0.48 & 0.84 & 11.0 & $128 \times 128$ \\ \bottomrule
    \end{tabular}
    }
    \caption{\textbf{Cost comparison.} Per scene cost comparison with Tri-Planes trained in the image space. Here, we consider $N_\mathrm{train}$ = 500, $N_\mathrm{exploit} = 1000$, $t_\mathrm{EC}$ = 40 hours, $M = 50$ , $F^\mathrm{mac} = 22$. Our method reduces the effective training time by 86\% per scene, and the effective memory cost by 44\% per scene.} 
    \label{tab:costs}
\end{table*}

\label{sec:exploiting-3Da-AE}
In this stage, we learn scenes using the 3D-aware latent space and the global Tri-Planes obtained in the former stage. Our exploitation dataset $\mathcal{X}_{\mathrm{exploit}}$ is composed of $N_\mathrm{exploit}$ scenes.
We start by randomly initializing our local Tri-Planes and training our scene representations with \textbf{Encode-Scene}.
Finally, to maximize the quality of the decoded RGB images, we finish this stage by a small fine-tuning of the decoder with an $\mathcal{L}_\mathrm{Dec}$ loss, similarly to \textbf{Decode-Scene}, but with a trainable decoder. \cref{alg:exploiting} details our exploitation procedure.

\section{Experiments}
\label{sec:experiments}
In this section, we present the resource costs and quality evaluations of our method.
We also present an ablation study to assess the added value of each element of our pipeline.

\paragraph{Dataset. }
We adopt the ShapeNet-Cars \citep{shapenet} dataset. 
Each scene $s$ is rendered at resolution $128 \times 128$ with 200 different camera poses, from which we take $90\%$ for training and $10\%$ for testing.
We use two subsets of scenes respectively for training our 3Da-AE, and exploiting it to learn numerous scenes.
For each scene, we divide the views into train and test views as to evaluate the NVS performances of our method.

\begin{figure}[t]
    \centering
    \includegraphics[width=\columnwidth]{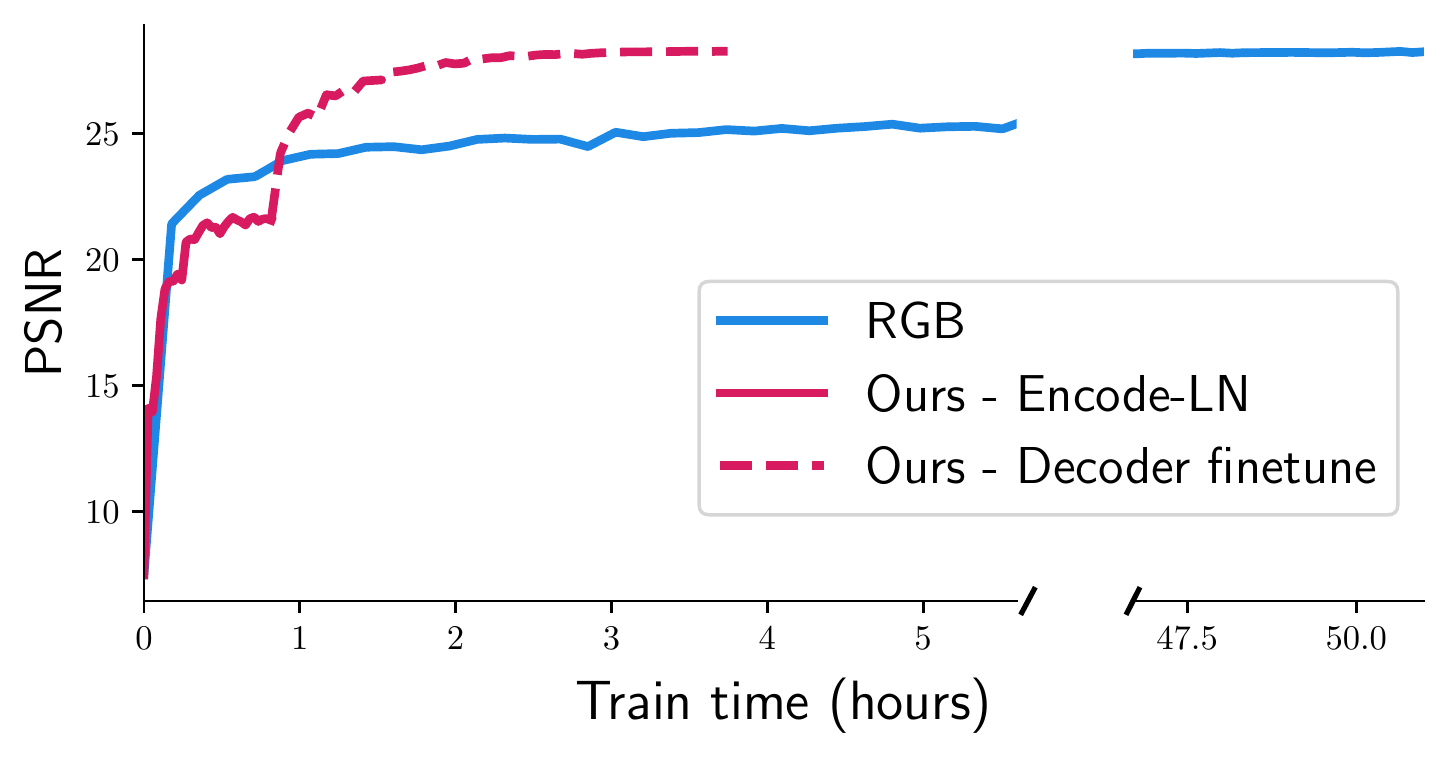}
    \caption{\textbf{Quality evolution.}  Evolution of the average test-view PSNR demonstrated in the exploit phase of our method compared to RGB Tri-Planes ($N_\mathrm{exploit} = 100$). Our method achieves comparable quality in less training time.}
    \label{fig:plot_ours_vs_rgb}
\end{figure}

\paragraph{Implementation details. }
We train the 3Da-AE on $N_\mathrm{train} = 500$ scenes from ShapeNet-Cars. For the exploitation phase, we learn $N_\mathrm{exploit} = 1000 $ scenes. For all our experiments, we take $F^\mathrm{mic} = 10$, $F^\mathrm{mac} = 22$, and $M = 50$. 

\subsection{Resource costs}
In this section, we detail the resource costs in terms of time and memory of the various stages of our method, and illustrate how we compare it to classical Tri-Plane training.

\begin{figure}[t]
    \centering
    \includegraphics[width=\columnwidth]{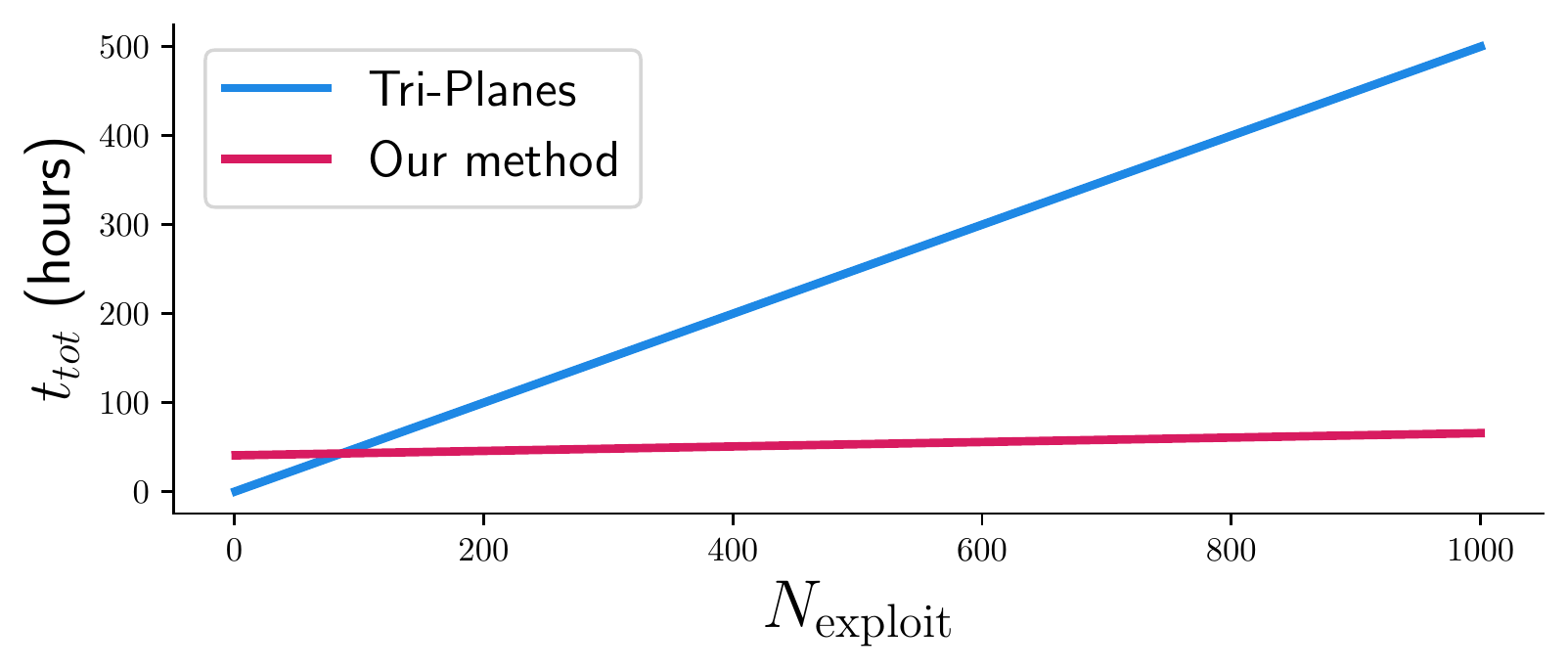}
    \caption{\textbf{Time cost evolution.} Total train time evolution when scaling the number of trained scenes $N_\mathrm{exploit}$. The entry training cost $t_{EC}$ is taken into account. Our method demonstrates more favorable scalability properties as compared to Tri-Planes (RGB).}
    \label{fig:memory_and_training_time_plots}
\end{figure}

\paragraph{Time costs. }
As presented, our method starts by warming up the Tri-Planes and training the 3Da-AE with $N_\mathrm{train}$ scenes, for which we respectively allocate the times $t_\mathrm{train}^\mathrm{warmup}$ and $t_\mathrm{train}$.
Subsequently, we exploit the 3Da-AE to train $N_\mathrm{exploit}$ scenes. 
We call $t_\mathrm{exploit}$ and $t_\mathrm{exploit}^\mathrm{ft}$ respectively the time to train the Tri-Planes in the exploit phase, and the time to fine-tune the decoder at the end of this phase.
Note that $t_\mathrm{train}^\mathrm{warmup}$ and $t_\mathrm{train}$ represent a time entry cost for our method, as this training is done only once, and it is independent of the number of scenes $N_\mathrm{exploit}$ which we wish to learn. We denote this entry cost time by $t_\mathrm{EC} = t_\mathrm{train}^\mathrm{warmup} + t_\mathrm{train}$~. Thus, our total training time is written as:

\begin{equation}
    t_\mathrm{tot} = t_\mathrm{EC} + N_\mathrm{exploit} t_\mathrm{scene}~,
\end{equation}
where $t_\mathrm{scene} = \frac{1}{N_\mathrm{exploit}} (t_\mathrm{exploit} + t_\mathrm{exploit}^\mathrm{ft})$ is the training time per scene in the exploit phase.
Finally, for a fair comparison to training in the RGB space, we define $t_\mathrm{scene}^\mathrm{eff}$, the effective time taken per scene in our method. This takes into account $t_\mathrm{EC}$, the entry cost to our method. $t_\mathrm{scene}^\mathrm{eff}$ is written as:
\begin{equation}
    t_\mathrm{scene}^\mathrm{eff} = \frac{t_\mathrm{EC}}{N_\mathrm{exploit}} + t_\mathrm{scene}~.
\end{equation}
Indeed, our method is more beneficial when $N_\mathrm{exploit}$ is large.

\begin{table}[t]
    \centering
    \footnotesize
    \resizebox{\columnwidth}{!}{
    \begin{tabular}{l c c c | c c c}
        \toprule
        Experiment & \rot{Latent Space} & \rot{Micro-Planes} & \rot{Macro-Planes} & \rot{Train scenes} & \rot{Exploit scenes} & \quad \\ \midrule
        Ours-Micro & \cmark & \cmark & \xmark & 26.52 & 26.95 & \\
        Ours-Macro & \cmark & \xmark & \cmark & 25.67 & 26.10 & \\
        Tri-Planes-Macro (RGB)& \xmark & \xmark & \cmark & 27.84 & 28.00& \\ \midrule
        Tri-Planes (RGB)& \xmark & \cmark & \xmark & \textbf{28.24} & 28.40 & \\
        Ours-No-Prior & \cmark & \cmark & \cmark & 27.72 & 28.13 & \\
        Ours & \cmark & \cmark & \cmark & 28.05 & \textbf{28.48} & \\ \bottomrule
    \end{tabular}
    }
    \caption{\textbf{Quality comparison.} Average PSNR demonstrated by our method with a comparison to Tri-Planes and ablations of our pipeline. All metrics are computed on never-seen test views. Here, we consider $N_\mathrm{train} = 500$, $N_\mathrm{exploit} = 100$, and $M = 50$. For compute constraints, Tri-Planes metrics are averaged on 50 scenes.}
    \label{tab:psnr_comp}
\end{table}
\paragraph{Memory costs. }
In terms of memory footprint, our method presents an advantage compared to its baseline as it requires less local Tri-Plane features per scene. 
We denote $m_E$, $m_D$ and $m_B$, the memory size required to respectively save the encoder, decoder, and the global planes. 
We also define $m_\mathrm{EC} = m_E + m_D + m_B$, the memory entry cost needed for our method.
Additionally, saving the scenes requires saving their local Tri-Planes of size $m_T$ and their coefficients of size $m_W$.
Thus, our total memory footprint is written as:
\begin{equation}
    m_\mathrm{tot} = m_\mathrm{EC} + N_\mathrm{exploit} m_\mathrm{scene}~,
\end{equation}
where $m_\mathrm{scene} = \frac{1}{N_\mathrm{exploit}} (m_T + m_W)$ is the memory size needed to save one scene.
Lastly, we define $m_\mathrm{scene}^\mathrm{eff}$, the effective memory size needed per scene, taking into account the entire pipeline. $m_\mathrm{scene}^\mathrm{eff}$ is written as:
\begin{equation}
    m_\mathrm{scene}^\mathrm{eff} = \frac{m_\mathrm{EC}}{N_\mathrm{exploit}} + m_\mathrm{scene}~.
\end{equation}
Similarly to the time costs, our method is also more advantageous with larger $N_\mathrm{exploit}$.

\subsection{Evaluations}
We apply our exploitation phase on two sets of scenes: scenes from the training set, and held-out scenes from the exploit set. 
We compare our results with a classical training of Tri-Planes in the image space, denoted \textbf{``Tri-Planes (RGB)''}. 
For a fair comparison, we use the same plane resolutions $T = 64$ and the same number of plane features $F = 32$ in all our experiments.
All results are obtained on never-seen test views belonging to scenes coming from both the train and exploit sets.
Note that, due to compute constraints, we only train Tri-Planes (RGB) on a reduced version of the datasets.

\begin{figure}[t]
    \centering
    \begin{subfigure}{0.32\columnwidth}
        \centering
        \includegraphics[width=0.8\textwidth]{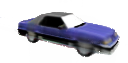}
    \end{subfigure}
    \begin{subfigure}{0.32\columnwidth}
        \centering
        \includegraphics[width=0.8\textwidth]{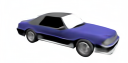}
    \end{subfigure}
    \begin{subfigure}{0.32\columnwidth}
        \centering
        \includegraphics[width=0.8\textwidth]{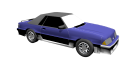}
    \end{subfigure}
    \\
    \begin{subfigure}{0.32\columnwidth}
        \centering
        \includegraphics[width=0.8\textwidth]{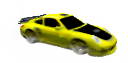}
        \caption{Tri-Planes (RGB)}
    \end{subfigure}
    \begin{subfigure}{0.32\columnwidth}
        \centering
        \includegraphics[width=0.8\textwidth]{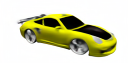}
        \caption{Ours}
    \end{subfigure}
    \begin{subfigure}{0.32\columnwidth}
        \centering
        \includegraphics[width=0.8\textwidth]{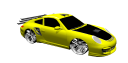}
        \caption{Ground truth}
    \end{subfigure}
    \caption{\textbf{Visual comparison}. Visual comparison of novel view synthesis quality for our method and Tri-Planes (RGB).}
    \label{fig:results}
\end{figure}

\paragraph{Results.}
As seen in \cref{tab:costs,tab:psnr_comp,fig:plot_ours_vs_rgb}, our method reach the same PSNR as Tri-Planes (RGB) while reducing the the training time by 86\% and the memory cost by 44\% when training $N_\mathrm{exploit}=1000$ scenes. 
In addition, rendering novel views using our latent Tri-Planes requires $53\%$ less time.
Qualitatively, \cref{fig:results} compares the novel view synthesis quality of our method with Tri-Planes (RGB) and ground truth test views.
Additionally, \cref{fig:memory_and_training_time_plots} shows how our method scales in  training time compared to Tri-Planes (RGB).

\subsection{Ablations}
To justify our choices and explore further, we present an ablation study of our method.
The first ablation, ``\textbf{Ours-Micro}'', eliminates the Micro-Macro decomposition, and consequently global information sharing (\ie $F^\mathrm{mac}=0~, F^{mic}=F$).
The second ablation, ``\textbf{Ours-Macro}'', eliminates local features from Tri-Planes and relies only on global features (\ie $F^{mic}=0~, F^\mathrm{mac}=F$).
The third ablation, ``\textbf{Tri-Planes-Macro}'' exclusively trains globally shared Tri-Planes in the image space instead of the latent space.
Finally, ``\textbf{Ours-No-Prior}'' refers to an ablation where we reset our globally shared Tri-Planes before the exploitation phase.
Note that ablating the latent space as well as information sharing is equivalent to the vanilla ``\textbf{Tri-Planes (RGB)}'' setting.
The results of our ablation study can be found in \cref{tab:psnr_comp}.

\section{Conclusion}
In this paper, we introduce a novel approach for efficiently learning abundantly many scenes. 
We propose a 3D-aware autoencoder that enables the training of scene representations in its latent space, drastically speeding-up rendering and training times.
Additionally, we present a Micro-Macro Tri-Planes scene decomposition enabling cross-scene information sharing and lighter scene representations. 
We show that our pipeline reduces resource costs required to learn an individual scene in both time and memory, while showing no quality loss.
We envision this work as an essential milestone towards a foundation 3D-aware latent space. 

{
    \small
    \bibliographystyle{ieeenat_fullname}
    \bibliography{main}
}

\end{document}